\documentclass[preprint,12pt]{elsarticle}

\usepackage{graphicx}
\usepackage{amssymb}
\usepackage{array}
\usepackage{multirow}
\usepackage{color}
\usepackage{amsmath}
\usepackage{ulem}
\usepackage{mathrsfs}
\usepackage{makecell}
\usepackage{subfigure}
\usepackage{soul}

\begin{document}
	
	\begin{frontmatter}
		
		
		\title{Question-Aware Memory Network for Multi-hop Question Answering in Human-Robot Interaction}
		
		
		\author{Xinmeng Li\textsuperscript{\rm 1}}
		\ead{xml.nudt@gmail.com}
		
		\author{Mamoun Alazab\textsuperscript{\rm 2}}
		\ead{mamoun.alazab@cdu.edu.an}
		
		\author{Qian Li\textsuperscript{\rm 1}}
		\ead{liqian9510@outlook.com}
		
		\author{Keping Yu\textsuperscript{\rm 3}}
		\ead{keping.yu@aoni.waseda.jp}
		
		\author{Quanjun Yin\textsuperscript{\rm 1}\corref{mycorrespondingauthor}}
		\ead{yinquanjun@nudt.edu.cn}
		
		\address{\textsuperscript{\rm 1}College of Systems Engineering, National University of Defense Technology, Changsha 410000, China}
		
		\address{\textsuperscript{\rm 2}College of Engineering, IT and Environment, Charles Darwin University, Australia}
		
		\address{\textsuperscript{\rm 3}Global Information and Telecommunication Institute, Waseda University, Tokyo, 169-0072 Japan}
		
		\cortext[mycorrespondingauthor]{Corresponding author.}

		
		
		
		\begin{abstract}
			Knowledge graph question answering is an important technology in intelligent human-robot interaction, which aims at automatically giving answer to human natural language question with the given knowledge graph. For the multi-relation question with higher variety and complexity, the tokens of the question have different priority for the triples selection in the reasoning steps. Most existing models take the question as a whole and ignore the priority information in it. To solve this problem, we propose question-aware memory network for multi-hop question answering, named QA2MN, to update the attention on question timely in the reasoning process. In addition, we incorporate graph context information into knowledge graph embedding model to increase the ability to represent entities and relations. We use it to initialize the QA2MN model and fine-tune it in the training process. We evaluate QA2MN on PathQuestion and WorldCup2014, two representative datasets for complex multi-hop question answering. The result demonstrates that QA2MN achieves state-of-the-art $Hits@1$ accuracy on the two datasets, which validates the effectiveness of our model.  
		\end{abstract}
		
		%
		%
		
	\end{frontmatter}
	
	
	\section{Introduction}
	\label{S:1}
	Intelligent human-robot interaction provides a convenient way for the communication between human and the robots~\cite{DBLP:journals/corr/abs-2101-00774,DBLP:journals/corr/abs-2101-09459,DBLP:conf/acl/KanHLJRY18,DBLP:conf/sigir/Lei0RC20}. Question answering over knowledge base (KBQA) is one of the important technology of intelligent human-robot interaction. It aims at using the given knowledge base to answer users' natural language question by cognitive computing~\cite{DBLP:journals/corr/abs-1907-09361}. The development of semantic web and the improvement of information acquisition technology promote the establishment and application of large-scale knowledge graph (KG), e.g. Freebase~\cite{bollacker2008freebase} , DBpedia~\cite{lehmann2015dbpedia}, etc. The massive information contained in knowledge graph further promotes the research and application of KBQA. Therefor, recent years have witnessed an increasing demand for conversational question answering agent that allows user to query a large-scale knowledge base (KB) in natural language~\cite{berant-etal-2013-semantic}. 
	
	
	It is a long-standing problem aims to answer user's natural language question using a structured knowledge base. A typical KB can be viewed as a knowledge graph consisting of entities, properties, and relations between them~\cite{oguz2020unified,DBLP:journals/corr/abs-2012-11957}. Historically, KBQA can be divided into two mainstreams \cite{chen-etal-2019-bidirectional}. The first branch, namely, the semantic parser method (SP-based method), tries to parse the natural language question into a logical form that can be used to query the knowledge base, e.g. SPARQL, $\lambda$-DCS \cite{liang2013lambda} and $\lambda$-calculus. 
	However, SP-based method heavily depends on data annotation and hand-crafted templates. The second branch treats KBQA as information retrieval problem, namely, information retrieval method (IR-based method). This approach encodes the question and each candidate as high-dimension vectors in a continuous semantic space and a ranking model is used to predict the correct answers. Recently, deep learning also leads an upward trend for IR-based methods. These approaches range from simple neural embedding based models \cite{bordes-etal-2014-question}, to attention based recurrent model \cite{hao-etal-2017-end}, then to memory-augmented neural controller architectures \cite{chen-etal-2019-bidirectional, DBLP:journals/corr/BordesUCW15,jain-2016-question}.
	
	\begin{figure}[!htbp]
		\centering
		\includegraphics[scale=0.8]{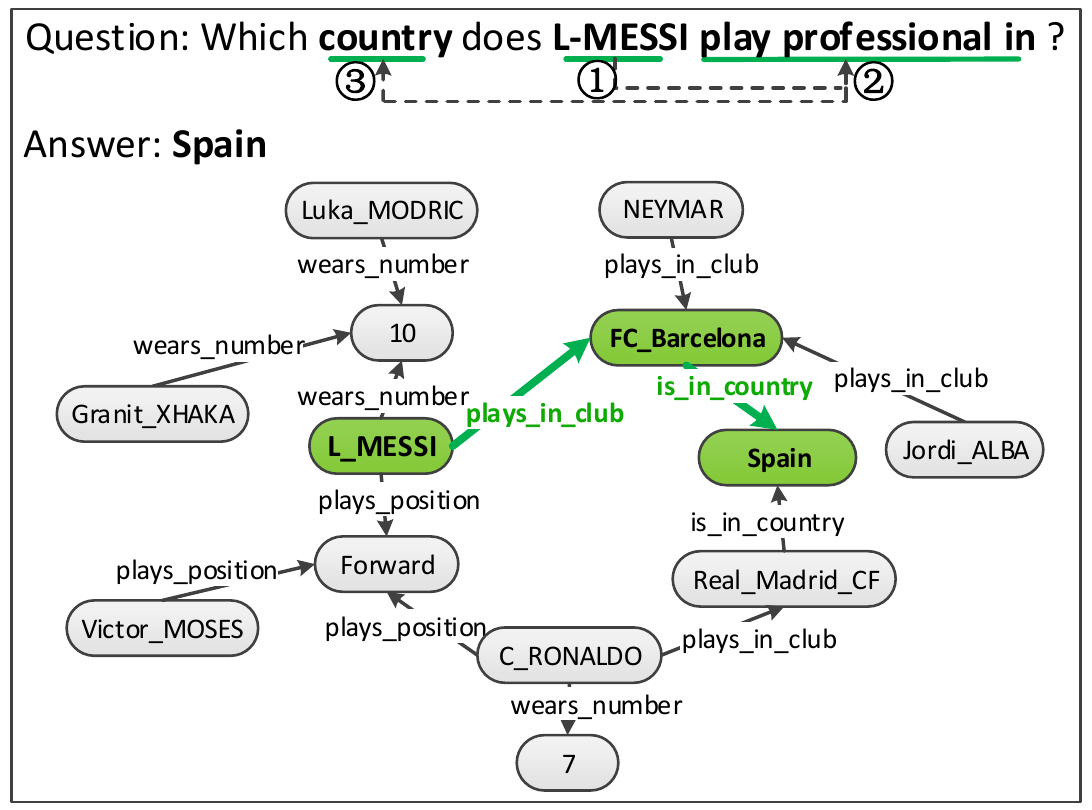}
		\caption{An example of multi-relation question over knowledge graph from WorldCup2014~\cite{DBLP:journals/corr/ZhangWT16}. The rounded rectangles represent the entities in KG and the solid arrows represent the relations between entities. The dot arrows represent the attention flow in the reasoning process. The entity ``L\_MESSI" is the first part to focus on, the phrase ``play professional in" next and ``country" finally.}
		\label{fig:example}
	\end{figure} 
	
	More recent work~\cite{zhou-etal-2018-interpretable, DBLP:conf/aaai/ZhangDKSS18, DBLP:journals/corr/abs-1908-06917,DBLP:journals/corr/abs-1808-10596} focuses on enhancing the reasoning capability for multi-hop question. \citet{DBLP:conf/kdd/LeiZ0MWCC20,DBLP:conf/wsdm/Lei0MWHKC20} proposed the multi-turn conversational recommendation series work under the human-computer interaction mode, which promoted the application and development of the human-computer interaction mode in NLP-related tasks such as recommendation and question answering. Specifically, multi-hop question means the question has multiple relations and needs more steps inference to get the final answer. For example in Figure~\ref{fig:example}, considering the question ``which country does L\_MESSI play professional in ?", where more than one relations (i.e., ``plays\_in\_club'' and ``is\_in\_country'') are involved.
	Due to the variety and complexity of knowledge and semantic information, multi-hop question answering over knowledge base is still a challenging task. 
	Generally, there are two challenges need to be addressed.
	
	First, the multi-hop question has more complicated semantic information. The tokens of the question have different influence on the triples selection in each reasoning step. 
	Take for example the question in Figure~\ref{fig:example}, The entity ``L\_MESSI" is the first part that should be focused on, the phrase ``play professional in" next and ``country" finally.	 
	Accordingly, the model should dynamically pay attention to different parts of the question during reasoning. However, current model often takes the question as a whole and ignore the priority information in it. 
	
	Second, the triplets have implicit relationship as some of them share entities or relations. From the way of humans thinking, we often find associated information from context. For example, ``FC\_Barcelona'' and ``Real\_Madrid\_CF'' share the same tail entity ``Spain'', which would enhance our memory that the two clubs are located in the same country. So the implicit graph context between triplets need to be modeled to improve the representation of entities and relations~\cite{nathani-etal-2019-learning}. However, previous work only considers the individual triplet and local information, the implicit graph context of knowledge base has not been fully explored.
	
	
	Considering the aforementioned challenges, we propose an architecture with question-aware attention to dynamically pay attention to different parts of the question in the reasoning process. We implement the architecture with key-value memory neural network, named \textbf{QA2MN} (\textbf{Q}uestion-\textbf{A}ware \textbf{M}emory \textbf{N}etwork for \textbf{Q}uestion \textbf{A}nswering), to update the attention on question timely during reasoning. To improve the representation of entities, we utilize KG embedding model to pre-train the embedding of entities and relations. For the triplets are modeled and scored independently in general KG model, we integrate graph context into the scoring function to enrich the semantic representation. 
	
	To summarize, we have three-fold contributions: (\romannumeral1) propose a novel architecture with question-aware attention in the reasoning process and implement it with QA2MN to improve the query update mechanism. (\romannumeral2) incorporate graph context information into KG embedding model to improve the representation of entities and relations. (\romannumeral3)  achieve state-of-the-art $Hits@1$ accuracy on two representative datasets and the ablation study demonstrates the interpretability of QA2MN.	
	
	The rest of the paper is structured as follows. We first give a review of related work in Section 2. Then background is showed in Section 3 and the detailed approaches are followed in Section 4. Experimental setups and results are reported in Section 5. Finally, we end the paper with conclusion and future work in Section 6.
	
	\section{Related work}
	\label{S:relatedwork}				
	Traditional SP-based models heavily depend on predefined templates instead of exploring the inherent information in knowledge graph \cite{Unger-1012-Template-Based, berant-etal-2013-semantic}. \citet{yih-etal-2015-semantic} proposes query graph method to effectively leverage the graph information by cutting the semantic parsing space and simplifies the difficulty of semantic matching. 
	For multi-hop question, \citet{xu-etal-2019-enhancing} uses key-value memory neural network to store the graph information, and a new query update mechanism is proposed to remove the key and value that has been located in the query when updating. So the model can better pay attention to the content that needs reasoning in the next step. The SP-based methods give logic form representation of natural language question and the query operation is followed to get the final answer. However, the SP-based methods more or less rely on feature engineering and data annotation. In addition, they are demanding for researchers to master the syntax and logic structures of data, which poses additional difficulties for non-expert researchers. 
	
	The IR-based methods treat KBQA as information retrieval problem by modeling questions and candidate answers with ranking algorithm. \citet{bordes-etal-2014-question} firstly employed embedding vectors to encode the question and knowledge graph into high-dimension semantic space. \citet{hao-etal-2017-end} presented a novel cross-attention based neural network model to consider the mutual influence between the representation of questions and the corresponding answer aspects, where attention mechanism was used to learn the dynamically relevance between answer and words in the question to effectively improve the matching performance.  \citet{chen-etal-2019-bidirectional} proposed bidirectional attentive memory network to capture the pairwise correlation between question and knowledge graph information and simultaneously improve the query expression by the attention mechanism. 
	However, those models are not enough to handle multi-relation questions due to the lack of multi-hop reasoning ability. 
	\citet{zhou-etal-2018-interpretable} proposed an interpretable, hop-by-hop reasoning process for multi-hop question answering. The model predicts the complete reasoning path till the final answer. However, considering the cost of data collection, it is scarcely possible to be generalized to other	domains. So weak-supervision\footnote{Full supervision means annotating the complete answer path till the final answer. The weak-supervision means only the final answer is labeled. The un-supervision means no label is needed. For example, considering the question ``which country does L\_MESSI play professional in ?", full supervision would annotate the complete answer path as (L\_MESSI, plays\_position, FC\_Barcelona), (FC\_Barcelona, is\_in\_country, Spain) and ``Spain'', while weak-supervision only resorts to the final answer ``Spain''. } with the final answer labeled is better suited to current needs.	
	The IR-based method converts the graph query operation into a data-driven learnable matching problem and can directly get the final answer by end-to-end training. Its advantages is that it reduces the dependence on hand-crafted templates and feature engineering, while the method is blamed for poor interpretability. 
	
	Recent work~\cite{DBLP:conf/aaai/ZhangDKSS18, qiu2020stepwise} also formulates multi-hop question answering as a sequential decision problem. \citet{DBLP:conf/aaai/ZhangDKSS18} treats the topic entity as a latent variable and handles multi-hop reasoning with variational inference. \citet{qiu2020stepwise} performs path search with weak supervision to retrieve the final answer. The model proposes a potential-based reward shaping strategy to alleviate the delayed and sparse reward problem. 
	
	
	\section{Background}
	\subsection{Task Description}
	\label{S:3}
	
	For the given structured knowledge graph $\mathscr{G}$, with entity set $\mathscr{E}$ and relation set $\mathscr{R}$, each triplet $T = (h, r, t) \in \mathscr{G}$ represents an atomic fact, where $h\in \mathscr{E}$, $t\in \mathscr{E}$, $r\in \mathscr{R}$ denote head entity, tail entity and the relaion between them. Given a natural language question $X$, the task is to reason over $\mathscr{G}$ and predict $Y$ to answer the question. Generally, the possible answers including (\romannumeral1) an entity from the entity set $\mathscr{E}$, (\romannumeral2) the numerical results of arithmetic operations, such as SUM or COUNT, and (\romannumeral3 ) one of the possible boolean values, such as True or False \cite{DBLP:journals/corr/abs-1907-09361}. In this paper, we mainly focus on the first problem of entity-centroid natural language question.	To facilitate understanding, we summarize the important symbols used in the paper in Table~\ref{tab:notations}.
	\begin{table*}[!ht]
		\centering
		\caption{\label{tab:notations} The important symbols and their definitions used in the paper.}
		\begin{tabular}{c|c}
			\hline
			Notations    & Definitions     \\ \hline  \hline
			$\mathscr{G}$ &  the knowledge graph \\ \hline
			$\mathscr{E}$ & the entity set   \\ \hline
			$\mathscr{R}$ & the relation set \\ \hline
			e & an entity in $\mathscr{E}$ \\ \hline			
			r & a relation in $\mathscr{R}$ \\ \hline
			h & the head entity in a triplet \\ \hline
			t & the tail entity in a triplet\\ \hline
			$(h, r, t)$ &  an atomic fact \\ \hline			
			$\psi(E_e,E_r,E_e)$ &  the scoring function   \\ \hline			
			X & a natural language question  \\ \hline
			x & the token in X \\ \hline
			$d_{ent}$ & the size of entity embedding\\ \hline
			$d_{rel}$ & the size of relation embedding\\ \hline
			$d_{emb}$ & the size of token embedding  \\ \hline
			$d_{hid}$ & the size of hidden representation \\ \hline
			$\Phi_V$ & the key embedding matrix  \\ \hline
			$\Phi_K$ & the value embedding matrix \\ \hline			
			$T_C$ & the candidate triplet set \\ \hline
			$A_C$ & the candidate answer set \\ \hline			
		\end{tabular}		
	\end{table*}
	\subsection{Preliminary}
	\subsubsection{KG Embedding}
	KG embedding converts symbolic representation of knowledge triples in a KG into continuous semantic spaces by embedding entities and relations into high-dimension vectors~\cite{KGE_A_Survey}. It can effectively improve the downstream tasks such as KG completion~\cite{bordes:hal-00920777,DBLP:journals/corr/abs-2010-07620}, relation extraction~\cite{weston:hal-00880455} and KBQA~\cite{saxena-etal-2020-improving}. 
	
	For each $e \in \mathscr{E}$ and $r \in \mathscr{R}$, KG embedding first maps it into continuous hidden representation $E_e$ and $E_r$. Then, a scoring function $\psi(E_e,E_r,E_e)$ assign a score to a possible triple $(h, r, t)$ to measure its plausibility. The triplets existed in $\mathscr{G}$ tend to have higher score than those not. To learn those entity and relation representations, an optimization method is used to maximize the total plausibility of observed Triplets.
	
	\subsubsection{Memory Neural Network}	
	\begin{figure}[htb]
		\centering 
		\subfigure[Memory neural network. $\Phi_A$ and $\Phi_C$ denote input embedding matrix and output embedding matrix.]
		{
			\begin{minipage}{0.35\linewidth}
				\flushleft
				\includegraphics[width=1\linewidth]{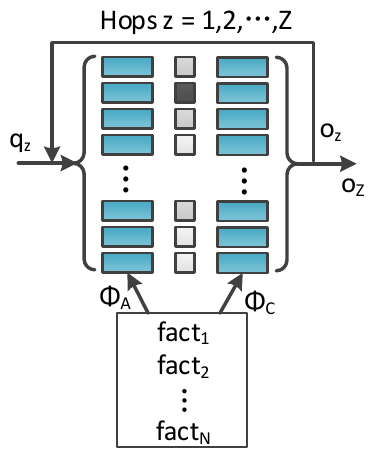}
			\end{minipage}
			\label{sub_fig:memorynetwork}
		}
		\subfigure[Key-value memory neural network.  $\Phi_K$ and $\Phi_V$ denote key embedding matrix and value embedding matrix.]
		{
			\begin{minipage}{0.38\linewidth}
				\flushright        
				\includegraphics[width=1\linewidth]{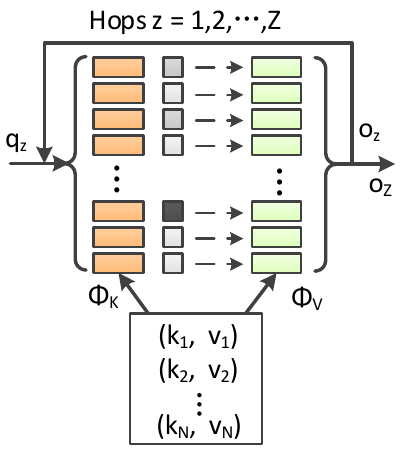}
			\end{minipage}
			\label{sub_fig:kv-memorynetwork}
		}
		\caption{The architecture of memory neural network and key-value memory neural network.}
		\label{fig:memory}
	\end{figure}
	
	The memory neural network \cite{DBLP:conf/nips/SukhbaatarSWF15} is well-known for its multiple hop reasoning ability and has been successfully applied in many natural language processing applications such as question answering \cite{chen-etal-2019-bidirectional} and reading comprehension \cite{DBLP:conf/nips/SukhbaatarSWF15}. A memory neural network is often stacked with multi-layers, each layer has two independent embedding matrices to transform the supporting facts into input memory representation and output memory representation. As shown in Figure~\ref{sub_fig:memorynetwork}, given the query vector, it first finds the supporting memories from the input memory representation and then produces output features by a weighted sum over the output memory representation. 
	
	Key-value memory neural network generalizes the standard memory network by dividing the memory arrays into two parts, i.e., the key slot and the value slot, as shown in Figure~\ref{sub_fig:kv-memorynetwork}. The model learns to use the query to address relevant memories with the keys, whose values are subsequently returned for output computation. Compared to the flat representation in standard memory network, the key-value architecture gives more flexibility to encode prior knowledge via functionality separation and is more applicable to complex structured knowledge sources \cite{xu-etal-2019-enhancing, miller-etal-2016-key}.
	
	\section{Proposed Model}
	\label{S:4}
	\begin{figure}[!htbp]
		\centering
		\includegraphics[width=1.0\linewidth]{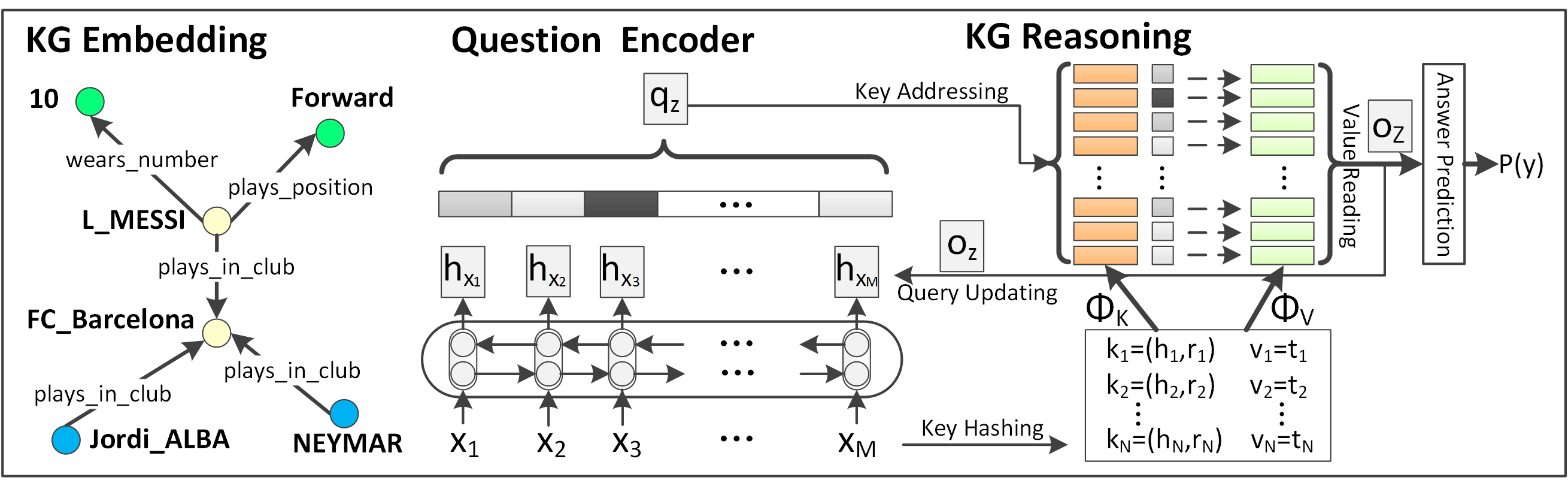}
		\caption{The architecture of QA2MN, which consists of three components (\romannumeral1) KG embedding, (\romannumeral2) Question Encoder and (\romannumeral3) KG reasoning, see Section~\ref{S:4} for details.  {\color{yellow}$\bullet$} denotes the head and tail entity of the concerned triplet, {\color{green}$\bullet$} denotes head-related context for ``L\_MESSI'', {\color{blue}$\bullet$} denotes tail-related context for ``FC\_Barcelona''.}
		\label{fig:architecture}
	\end{figure}
	
	We use a three-stage model for question answering.  First, we exploit the graph context information in knowledge base by pre-training KG embedding model. Then, we use Bi-directional Gated Recurrent Unit ($\text{BiGRU}$) to encode the question into continuous hidden representation. Finally, we using a question-aware key-value memory network to reason over the knowledge graph. The proposed QA2MN has three main components, i.e., KG Embedding, Question Encoder and KG Reasoning. Figure~\ref{fig:architecture} illustrates the architecture.
	
	\subsection{KG Embedding with Graph Context}
	\label{S:4.1}
	
	We adopt translational distance model~\cite{Lin-2015-Learning} to train the embedding of entities and relations. For each fact $T_i = (h_i, r_i, t_i) \in \mathscr{G}$, we apply translational distance constraint for the entities and the relations by the following equation, 
	\begin{equation}
	W_{e2r}E_{h_i} + E_{r_i} = W_{e2r}E_{t_i}
	\end{equation}
	where $E_{h_i}\in \mathbf{R}^{d_{ent}}$, $E_{r_i}\in \mathbf{R}^{d_{rel}}$ and $E_{t_i}\in \mathbf{R}^{d_{ent}}$ are the embeddings of head entity, relation, and tail entity respectively, $W_{e2r}\in \mathbf{R}^{d_{rel}\times d_{ent}}$ is a projection matrix from the entity space to the relation space. In our implementation, $d_{ent}$ is equal to $d_{rel}$. 
	Then, we obtain the translational distance score by,
	\begin{equation}
	\label{eq:dt}
	d_i^{t} = \|W_{e2r}E_{h_i} + E_{r_i} - W_{e2r}E_{t_i}\|
	\end{equation}
	where $\|\cdot\|$ denotes the $l_2$ norm of variable.
	To explore the implicit context information of knowledge graph, we integrate graph context into the distance scoring to improve the representation of entities.	
	For the triplet $T_i$, we consider two kinds of context information: (\romannumeral1) head-related context: all the triples share the same head with $T_i$, i.e., $C_h(T_i) = \{T_j|T_j=(h_j, r_j, t_j) \in G, h_j=h_i\}$; (\romannumeral2) tail-related context: all the triples share the same tail with $T_i$, i.e., $C_t(T_i) = \{T_j|T_j=(h_j, r_j, t_j) \in G, t_j=t_i\}$.
	
	First, we integrate the head-related context with $E_{h_i}$ by taking average over the triplets from $C_h(T_i)$,
	\begin{equation}
	\begin{aligned}
	\tilde{E}_{h_i}&=\frac{\sum_{((h_j,r_j, t_j)\in C_h(T_i)}\hat{E}_{h_j}}{|C_h(T_i)|} \\
	\hat{E}_{h_j} &=  E_{t_j} - W_{e2r}^{-1} E_{r_j}
	\end{aligned}
	\end{equation}
	where, $|C_h(T_i)|$ is the number of head-related context triplets, $-1$ is the inverse operator. Then, we compute the distance of the head-related context representation and $E_{h_i}$ by,
	\begin{equation}
	\label{eq:dch}
	d_i^{C_h} = \|E_{h_i}-\tilde{E}_{h_i}\|
	\end{equation}
	
	In the same way, we compute the tail-related context representation $\tilde{E}_{t_i}$ as the average from triplets in $C_t(T_i)$,
	\begin{equation}
	\begin{aligned}
	\tilde{E}_{t_i} &= \frac{\sum_{((h_j,r_j, t_j)\in C_t(T_i)}\hat{E}_{t_j}}{|C_t(T_i)|} \\
	\hat{E}_{t_j} &=  E_{h_j} + W_{e2r}^{-1} E_{r_j}
	\end{aligned}
	\end{equation}	
	where $|C_t(T_i)|$ is the number of tail-related context triplets. Correspondingly, the distance of the tail-related context representation and $E_{t_i}$ is computed by,
	\begin{equation}
	\label{eq:dct}
	d_i^{C_t} = \|E_{t_i}-\tilde{E}_{t_i}\|
	\end{equation}
	
	\subsection{Question Encoder}
	\label{S:4.2}
	We use $\text{BiGRU}$~\cite{chung2014empirical} to encode the question to keep the token-level and sequence-level information. With a question $X = [x_{1},x_{2},...,x_M]$, where $M$ is the total number of tokens in $X$. We feed $X$ into the $\text{BiGRU}$ encoder, which is computed as follows,
	\begin{equation}
	\begin{aligned}
	\overrightarrow{h}_{x_i} = \text{GRU}(E_{x_i}, \overrightarrow{h}_{x_{i-1}}) \\		
	\overleftarrow{h}_{x_i} = \text{GRU}(E_{x_i}, \overleftarrow{h}_{x_{i+1}})
	\end{aligned}		
	\end{equation}
	where $\text{GRU}$ is the standard Gated Recurrent Unit, $E_{x_i} \in \mathbf{R}^{d_{emb}}$ is the embedding of token, $\overrightarrow{h}_{x_i}\in \mathbf{R}^{d_{hid}/2},\overleftarrow{h}_{x_i}\in \mathbf{R}^{d_{hid}/2}$ is the hidden representation, $d_{emb}$ is the token embedding size and $d_{hid}$ is the hidden size. Then we obtain the hidden representations for each token, $H_{X} = [h_{x_1},h_{x_2},...,h_{x_M}]$, where $h_{x_i}$ is the concatenation of $\overrightarrow{h}_{x_i}$ and $\overleftarrow{h}_{x_i}$, i.e., $h_i=[\overrightarrow{h}_{x_i},\overleftarrow{h}_{x_i}]$.
	
	\subsection{KG Reasoning}
	\label{S:4.3}
	The focus of vanilla key-value memory neural network is about understanding the knowledgeable triplets in the memory slots. It often encodes the question as a whole vector and ignores its priority information. It is relatively enough for single-relation question but insufficient for complex multi-hop question. To improve the reasoning ability of key-value memory neural network, we introduce QA2MN to dynamically pay attention to different parts of the question in each reasoning step. In implement, QA2MN consists of five parts, i.e., key hashing, key addressing, value reading, query updating and answer prediction. 
	\subsubsection{Key Hashing}
	Key hashing uses the question to select a list of candidate triplets to fill the memory slot. Specifically, we first detect core entity as the entity mentioned in the question and find out its neighboring entities within $K$ hops relation. Then, we extract all triplets in $\mathscr{G}$ that contains any one of those core entities as the candidate triplets, denoted as $T_C = \{T_1,T_2,...,T_N\}$, where $N$ is the number of candidate triplets. All the entities in $T_C$ are extracted as candidate answers, we denote it as $A_C = \{A_1,A_2,...,A_L\}$, where $L$ is the number of candidate answers. For each candidate triplet $T_i = (h_i, r_i, t_i) \in T_C$, we store the head and relation in the $i$-th key slot, which is denoted as,
	\begin{equation}
	\Phi_K(k_i) = W_k(W_{e2r}E_{h_i} + E_{r_i})
	\end{equation}
	
	Correspondingly, the tail is stored in the $i$-th value slot, denoted as,
	\begin{equation}
	\Phi_V(v_i) = W_v E_{t_i}
	\end{equation}
	where $W_k \in \mathbf{R}^{d_{hid}\times d_{rel}}$ and $W_v \in \mathbf{R}^{d_{hid}\times d_{ent}}$ are trainable parameters. 
	
	At the $z$-th reasoning hop, QA2MN makes multiple hop reasoning over the memory slot by (\romannumeral1) computing relevance probability between query vector $q_z \in \mathbf{R}^{d_{hid}}$ and the key slots, (\romannumeral2) reading from the value slots, (\romannumeral3) updating the query representation based on the value reading output and the question hidden representation.
	
	\subsubsection{Key Addressing}
	Key addressing computes the relevance probability distribution between $q_z$ and $\Phi_K(k_i)$ in the key slots,
	\begin{equation}
	p_i^{qk} = \text{softmax}(q_z \Phi_K(k_i))
	\end{equation}
	\subsubsection{Value Reading}
	Value reading component reads out the value of each value slot by taking the weighted sum over them with $p_i^{qk}$,
	\begin{equation}
	o_z = \sum_{i=1}^{N} p_i^{qk} \Phi_V(v_i)
	\end{equation}
	\subsubsection{Query Updating}
	The value reading output is used to update the query representation to change the query focus for next hop reasoning. First, we compute the attention distribution between the value reading output $o_z$ and the hidden representation of each token in the question,
	\begin{equation}
	p_i^{vq} = \text{softmax}(o_z h_{x_i})
	\end{equation}
	
	Then, we update the query vector by summing the value reading output $o_z$ and the weighted sum over tokens in question with $p_i^{vq}$:
	\begin{equation}
	q_{z+1} = o_z + \sum_{i=1}^{M} p_i^{vq} h_{x_i}
	\end{equation}
	\subsubsection{Answer Prediction}
	We initialize the query $q_1$ with the self attention of the question representation, 
	\begin{equation}
	q_1 = \sum_{i=1}^{M}{\text{softmax}(h_x^\text{T}h_{x_i})h_{x_i}}
	\end{equation}
	where, $h_x=[\overrightarrow{h}_{x_M},\overleftarrow{h}_{x_1}]$ is the integrated representation of question, $\text{T}$ is the transposition operator.
	After $Z$ hops of reasoning over the memories, the final value representation $o_Z$ is used to perform the final prediction over all candidate answers. Finally, we compute the matching score between final value representation $o_Z$ and candidate answers and normalize it into the range of (0,1), 
	\begin{equation}
	P(y) = \text{softmax}(W_p o_{Z})
	\end{equation}
	where $W_p \in \mathbb{R}^{L\times d_{hid}}$ is a trainable parameter. Finally, the candidate answers are ranked by their score.
	
	\subsection{Training}
	\label{S:4.4}
	The training process can be divided into two stages. We first pre-train the KG embedding for several epochs, then we optimize the parameters of QA2MN and KG embedding iteratively. We combine the three distance score stated in Equation (\ref{eq:dt},\ref{eq:dch},\ref{eq:dct}) as the loss function for KG embedding training,
	\begin{equation}
	L_{KGE} = \sum_{T_i \in G}{d_i^{t} + d_i^{C_h} + d_i^{C_t}}
	\end{equation}
	
	As for QA2MN optimization, we use the cross-entropy to define the loss function. Given an input question $X$, we denote $y$ as the gold answer and $\tilde{y}$ as the predicted answer distribution. We compute the cross-entropy loss between $y$ and $P(y)$ by,
	\begin{equation}
	L_{QA} = -\sum_{X}{y\cdot log P(y)}
	\end{equation}
	
	\section{Experiments}
	\label{S:5}
	\subsection{Dataset}
	PathQuestion \cite{zhou-etal-2018-interpretable} and WorldCup2014 \cite{DBLP:journals/corr/ZhangWT16} are two representative datasets for complex multi-hop question answering, we employ them to evaluate QA2MN and the baselines. 
	
	\begin{table*}[!htbp]
		\centering
		\caption{\label{tab:datasetstatistics} Dataset statistics of PQ, PQL and WC. `\#' means number of items and $|$X$|$ represents the average length of questions.}
		\setlength{\tabcolsep}{3mm}{
			\begin{tabular}{|l|c|c|c|c|c|c|c|c|}
				\hline 
				\multirow{2}{*}{Dataset} &\multicolumn{2}{c|}{PQ} & \multicolumn{2}{c|}{PQL} &  \multicolumn{4}{c|}{WC} \\
				\cline{2-9}
				& 2H   &3H    &2H   & 3H  & 1H   & 2H  & M   &  C     \\ \hline \hline
				\#Entity   &1057  &1837  &5027 &6497 & \multicolumn{4}{c|}{1128} \\ \hline
				\#Relation &14    &14    &364  &412  & \multicolumn{4}{c|}{11}   \\ \hline
				\#Triplet  &1211  &2839  &4247 &5597 & \multicolumn{4}{c|}{6482} \\ \hline			
				\#Question &1908  &5198  &1594 &1031 &6482  &1472 &7954 &2208    \\ \hline
				\makecell[c]{$|$X$|$}    &8.1   &10.7  &9.0  &11.0 &6.84  &9.8  &7.4  &9.52    \\ \hline
		\end{tabular}}		
	\end{table*}

	\noindent $\bullet$ PathQuestion (PQ): It is a manually generated dataset with predefined templates and its knowledge base is adopted from subset of FB13 \cite{DBLP:conf/nips/SocherCMN13}. PathQuestion-Large (PQL) is more challenging with less training instances and larger scale of knowledge base adopted from Freebase \cite{bollacker2008freebase}. Both contain two-hop relation questions (2H) and three-hop relation questions (3H). 
	
	\noindent $\bullet$ WorldCup2014 (WC): The dataset is based on the knowledge base about soccer players that participated in FIFA World Cup 2014. It contains single-relation questions (1H), two-hop relation questions (2H), and conjunctive questions (C); M denotes the mixture of 1H and 2H. The statistics of PathQuestion and WorldCup2014 are listed in Table~\ref{tab:datasetstatistics}. 
	
	The complete KG setting in original dataset is too ideal for the question because there is often missing link in practical application. So the model should also be able to work on an incomplete KG setting. Following \citet{zhou-etal-2018-interpretable}, we simulate an incomplete KG setting, named PQ-50, by randomly removing half of the triples from the PQ-2H dataset.
	
	\subsection{Evaluation Metric}
	Following \citet{qiu2020stepwise}, we measure the performance of models by $Hits@1$, which is the percentage of examples the predicted answer exactly matches the gold one. When a question has multiple possible answers, the predicted answer would be correct if matching any one of them.
	
	\subsection{Implementation Detail}
	For the training of QA2MN, we use ADAM \cite{kingma2015adam} to optimize the trainable parameters. Gradients are clipped when their norm is bigger than 10. We partition the datasets in the proportion of 8:1:1 for training, validating and testing. The batch size is set to 48. The relation hop $K$ is set to 3 and the reasoning hop $Z$ is set to 3. The learning rate is initialized to $10^{-3}$ and exponentially annealed in the range of [$10^{-3}$, $10^{-5}$] with a decay rate of 0.96. The entity embedding dimension and the relation embedding dimension are set to 100. The token embedding dimension and hidden size are also set to 100. To increase model generalization, dropout mechanism is adopted by randomly masking 10\% of the memory slots.
		
		For the pre-training of KG embedding, we set the same optimizer and embedding dimension as above. We set the batch size to 64 and pre-train KG embeddings for 20 epochs.
		\subsection{Baseline}
		We have six baselines for comparison, including current state-of-the-art model. All of them are listed as follow,
		
		\noindent $\bullet$ Seq2Seq \cite{DBLP:journals/corr/SutskeverVL14}. It is an encoder-decoder model, adopting a LSTM to encode the input question sequence and another LSTM to decode the answer path.
		
		\noindent $\bullet$ MemNN \cite{DBLP:conf/nips/SukhbaatarSWF15}. It is an end-to-end memory network that stores the KG triplets in memory arrays by bag-of-words representation.
		
		\noindent $\bullet$ KV-MemNN \cite{miller-etal-2016-key}. It uses a key-value memory neural network to generalize the original memory network by dividing the memory arrays into two parts. For each triplet, the head and the relation are stored in the key slot, and the tail is stored in the value slot. 
		
		\noindent $\bullet$ IRN \cite{zhou-etal-2018-interpretable}. It proposes an interpretable, hop-by-hop reasoning process to predict the complete intermediate relation path. The answer module chooses the corresponding entity from KB at each hop and the last selected entity is chosen as the answer.
		
		\noindent $\bullet$ IRN-weak. IRN needs label the complete paths from topic entities to gold answers, which need extra annotation for the dataset. IRN-weak is a variant of IRN which only utilizes supervision from the final answer.
		
		\noindent $\bullet$ SRN \cite{qiu2020stepwise}. SRN formulates multi-relation question answering as a sequential decision problem. The model performs path search over the knowledge graph to obtain the answer and proposes a potential-based reward shaping strategy to alleviate the delayed and sparse reward problem caused by weak supervision.  
		
		\subsection{Experimental Result}
		
		\begin{table*}[!htbp]
			\centering
			\caption{\label{tab:result}  $Hits@1$ accuracy of QA2MN and the baselines on the two datasets. This table is split into two parts: full-supervised method on the upper side and weak-supervised method on the bottom side. }
			\setlength{\tabcolsep}{0.6mm}{
				\begin{tabular}{|l|c|c|c|c|c|c|c|c|c|}
					\hline 
					\multirow{2}{*}{Model} & \multicolumn{3}{c|}{PQ} & \multicolumn{2}{c|}{PQL}& \multicolumn{4}{c|}{WC} \\
					\cline{2-10}
					& 2H    & 3H & -50 & 2H    & 3H    & 1H    & 2H    & M     & C     \\ \hline \hline
					Seq2Seq        &0.899  & 0.770 & -     & 0.719 & 0.647 & 0.537 & 0.548 & 0.538 & 0.577 \\ \hline		
					MemNN          & 0.930 & 0.845 & 0.899 & 0.690 & 0.617 & 0.854 & 0.915 & 0.907 & 0.733  \\ \hline
					KV-MemNN       & 0.937 & 0.879 & 0.902 & 0.722 & 0.674 & 0.870 & 0.928 & 0.905 & 0.788 \\ \hline
					IRN            & 0.960 & 0.877 & 0.937 & 0.725 & 0.710 & 0.843 & 0.981 & 0.907 & 0.910 \\ \hline \hline
					IRN-weak       & 0.919 & 0.833 & -     & 0.630 & 0.618 & 0.749 & 0.921 & 0.786 & 0.837 \\ \hline
					SRN            &\textbf{0.963}
					& 0.892 & -     & 0.786 & 0.775 &\textbf{ 0.989} 
					& 0.978 & 0.965 & 0.873 \\ \hline
					\textbf{QA2MN} & 0.958 &\textbf{0.914} 
					& \textbf{0.944} 
					& \textbf{0.849} 
					&\textbf{0.853} 
					& 0.986 & \textbf{0.981} 
					& \textbf{0.971}
					& \textbf{0.923}\\ \hline
			\end{tabular}}		
		\end{table*}
		
		The results are shown in Table~\ref{tab:result}. QA2MN outperforms or shows comparable performance to all the baselines on the two datasets, which demonstrates that QA2MN is effective and robust in face with different datasets and questions. 	
		Seq2Seq shows the worst performance on the two datasets, indicating that multi-hop question answering is a challenging problem and the vanilla Seq2Seq model is not good at the complex reasoning process.	
			KV-MemNN always outperforms MemNN, confirming that the key-value architecture of KV-MemNN gives more flexibility to encode the triplet in KG and is more applicable to the multi-hop reasoning problem.	
		After further observations, we draw the following conclusions, 
		
		(1) QA2MN shows robustness on both simple and complex question.	
		
		We classify the simple question as dataset with less hops and larger data scale, including PQ-2H and WC-1H. Correspondingly, complex question has more hops and smaller data scale, including PQ-3H, PQL-3H, WC-2H and WC-C.
		
		As can be seen from Table~\ref{tab:result}, QA2MN performs similar to prior state-of-the-art model in case of simple question, since it has less challenge to predict the correct answer as the answer is directly connected to the core entity. For the complex question, IRN and SRN significantly lag behind QA2MN, showing that multi-hop reasoning is also challenging to prior state-of-the-art model. IRN initializes the question by adding the token embeddings as a whole vector, which would loss the priority information in question. The action space of SRN would exponentially growth as the reasoning hop increasing. So the performance drop is unavoidable for IRN and SRN as the question becomes more complex. On the other hand, the highest score on PQ-3H, PQL-3H, WC-2H and WC-C reveals QA2MN is able to precisely focus on the proper position of the question and infer the correct entity from the candidate triplets. So the result suggests that QA2MN is more robust when facing with complex multi-hop questions.
		
		(2) QA2MN is effective on incomplete KG setting.
		
		In the incomplete KG setting, only half of the original triples are reserved. Current model like IRN requires a path between the core entity and the answer entity. On the other hand, QA2MN uses dropout mechanism to randomly mask triplets in the memory slot to prevent it from over-fitting. QA2MN can implicitly explore the observed and unobserved paths around the core entity, which greatly improve the robustness of model to deal with the incomplete setting. So even there is no path between the core and answer entity, QA2MN can work to predict the answer.	
		
		(3) QA2MN meets current demand with weak-supervision learning.
		
		IRN outperforms IRN-weak for IRN need full-supervision along the whole intermediate relation and entity path. However, full-supervised method need large amount of data annotation which is cost and impractical for most case \cite{qiu2020stepwise}. That is to say, weak-supervised or unsupervised method is more suitable to the current demand.	
		
		QA2MN and SRN achieved the best and second-best performance on the two datasets, which confirms that weak-supervised method has great potential to explore the inherent semantic information in knowledge graph.
		
		\subsection{Ablation Study}
		
		To further verify the significance of the question-aware query update mechanism and knowledge graph embedding, we do model ablation to explore the following two questions: (\romannumeral1) is KG embedding necessary for model training? (\romannumeral2) is the question-aware query update mechanism helpful for reasoning over the knowledge graph? We use two ablated models to answer them.
		
		\noindent $\bullet$ QA2MN$\backslash$KE. It removes the pre-training of KG embedding .
		
		\noindent $\bullet$ QA2MN$\backslash$QA. It removes the question-aware query update mechanism and replaces it with standard key-value memory neural network.
		
		\begin{table*}[!htbp]
			\centering
			\caption{\label{tab:ablation} $Hits@1$ accuracy of ablation models on PQ and PQL.}
			\setlength{\tabcolsep}{4mm}{
				\begin{tabular}{|l|c|c|c|c|}
					\hline 
					\multirow{2}{*}{Model} & \multicolumn{2}{c|}{PQ} & \multicolumn{2}{c|}{PQL} \\
					\cline{2-5}
					& 2H    & 3H    & 2H    & 3H    \\ \hline \hline
					QA2MN               &0.958  &0.914  &0.944  &0.849  \\ \hline
					QA2MN$\backslash$KE &0.942  &0.889  &0.910  &0.819  \\ \hline
					QA2MN$\backslash$QA &0.939  &0.902  &0.884  &0.765  \\ \hline
					KV-MemNN            &0.937  &0.879  &0.722  &0.674  \\ \hline
			\end{tabular}}		
		\end{table*}
		
		We evaluate the ablation models on PQ and PQL dataset and take KV-MemNN for comparison. As shown in Table~\ref{tab:ablation}, comparing with QA2MN, the performance obviously dropped after removing any one of the two components, which confirms that both the question-aware query update mechanism and knowledge graph embedding are effective for improving the model performance.
		
		QA2MN$\backslash$QA always outperforms KV-MemNN, which approves that KG embedding adds context information from knowledge base to improve the representation of entities and relations.	
		QA2MN$\backslash$KE outperforms KV-MemNN as well, which confirms that the question-aware query update mechanism could improve the model to deal with more complex questions. To account for the performance improvement, we visualize the weight distributions on the question during the reasoning process in next subsection. 
		
		\subsection{Visualization Analysis}
		To illustrate how QA2MN allocates the attention hop-by-hop in the reasoning process, we choose a testing example from PathQuestion and visualize the attention distributions on the question in each reasoning step.
		
		\begin{figure}[!htbp]
			\centering
			\includegraphics[scale=1.5]{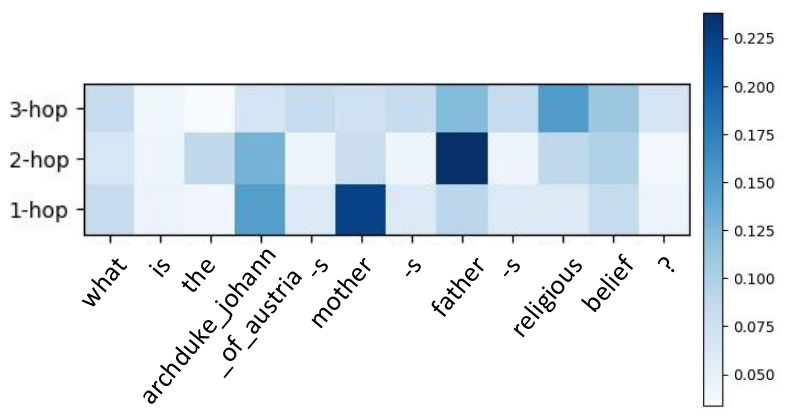}
			\caption{Attention weight heat-map of question ``what is the archduke\_johann\_of\_austria -s mother -s father -s religious belief ?''. The columns are the tokens in question and rows are the attention weight in each reasoning step.}
			\label{fig:heatmap}
		\end{figure}
		
		Figure~\ref{fig:heatmap} shows the attention heat-map of question ``what is the archduke\_johann\_of\_austria -s mother -s father -s religious belief ?''. The question contains a core entitiy (i.e., ``archduke\_johann\_of\_austria'') and three relations ( i.e., ``mother'', ``father'' and ``religious belief''). To answer the question, three triplets, i.e., (archduke\_johann\_of\_austria, parents, maria\_louisa\_of\_spain), (maria\_louisa\_of\_spain, parents, charles\_iii\_of\_spain) and (charles\_iii\_of\_spain, religion, catholicism) are needed to enable the reasoning. From Figure~\ref{fig:heatmap}, we find that QA2MN can focus on the correct position during reasoning process as human do. The question-aware attention detects relation ``mother'' initially. Then the attention turn to ``father'' and focuses on ``religious belief'' finally.  
		
		Previous work often uses bag-of-word representation or RNN/LSTM/GRU to encode the question into an integrated vector, resulting in the loss of inherent priority information in the sentence. In the reasoning process, the integrated vector is used to retrieve and rank the candidate triplets. It is challenging for the coarse-grained semantic representation to do complex reasoning.	Figure~\ref{fig:heatmap} intuitively illustrates the fine-grained information brought from question-aware attention, which is also the main reason for performance improvement. That is to say, question-aware attention can effectively explore the priority of the question, and utilize the fine-grained information for precisely reasoning.

		\section{Conclusion}
		\label{S:6}
		Multi-hop question answering over knowledge bases is a challenging task. There are two main aspects need to be addressed. First, multi-hop questions have more various and complicated semantic information. Then, the triplets have implicit relation as some of them share the heads or tails. We propose QA2MN to dynamically focus on different parts of the questions during reasoning steps. In addition, KG embedding is incorporated to learn the representation of entities and relations to extract the context information in knowledge graph. Extensive experiments demonstrate that QA2MN achieves state-of-the-art performance on two representative datasets. 
		
		In application, there are more complex questions which need arithmetic function or boolean logical operation. Furthermore, user may ask sequential questions continuously, which would lead to co-reference resolution problem. We would explore these problems in future work. 
		\\
		\\
		
		
		
		
		
		
		\noindent\textbf{Conflict of interest statement}
		\\
		
		On behalf of all authors, the corresponding author states that there is no conflict of interest.
		\\
		\\
		\noindent\textbf{REFERENCES}
		\bibliographystyle{elsarticle-num-names}
		\bibliography{sample}
		
		
		
		
		
		

	\end{document}